\documentclass[letterpaper]{article} % DO NOT CHANGE THIS
\usepackage{aaai23}  % DO NOT CHANGE THIS
\usepackage{times}  % DO NOT CHANGE THIS
\usepackage{helvet}  % DO NOT CHANGE THIS
\usepackage{courier}  % DO NOT CHANGE THIS
\usepackage[hyphens]{url}  % DO NOT CHANGE THIS
\usepackage{graphicx} % DO NOT CHANGE THIS
\usepackage{graphicx} % DO NOT CHANGE THIS
\usepackage{natbib}  % DO NOT CHANGE THIS AND DO NOT ADD ANY OPTIONS TO IT
\usepackage{caption} % DO NOT CHANGE THIS AND DO NOT ADD ANY OPTIONS TO IT
\usepackage{newfloat}
\usepackage{listings}
\DeclareCaptionStyle{ruled}{labelfont=normalfont,labelsep=colon,strut=off} % DO NOT CHANGE THIS
\title{`` It's a Match! '' \\ A Benchmark of Task Affinity Scores for Joint Learning}

\author{
    Raphaël Azorin,\textsuperscript{\rm 1}
    Massimo Gallo,\textsuperscript{\rm 1}
    Alessandro Finamore, \textsuperscript{\rm 1}
    Dario Rossi,\textsuperscript{\rm 1}
    Pietro Michiardi \textsuperscript{\rm 2}
}
    
\affiliations{
  \textsuperscript{\rm 1}{Huawei Research Center, France}\\
  \textsuperscript{\rm 2}{Eurecom, France}\\
  first.last@huawei.com,
  first.last@eurecom.fr
}

\newcommand{\eg}{{e.g.,}} 
\newcommand{\ie}{i.e.,} 

\usepackage[dvipsnames]{xcolor}
\usepackage{mathtools}
\usepackage{bm}
\usepackage{tablefootnote}
\usepackage{tabularx}
\usepackage{makecell}
\usepackage{amsmath}
\usepackage{ragged2e}
\usepackage{ifthen}
\usepackage{tikz}
\usepackage{collcell}
\usepackage{transparent}
\usepackage[dvipsnames]{xcolor}
\usepackage{mathtools}
\usepackage{bm}
\usepackage{tablefootnote}
\usepackage{tabularx}
\usepackage{makecell}
\usepackage{amsmath}
\usepackage{ragged2e}
\usepackage{ifthen}
\usepackage{tikz}
\usepackage{collcell}
\usepackage{transparent}
\usepackage{appendix}

\usepackage[
pdfauthor={Raphael Azorin, Massimo Gallo, Alessandro Finamore, Dario Rossi, Pietro Michiardi},
pdftitle={"It’s a Match!” A Benchmark of Task Affinity Scores for Joint Learning},
pdfkeywords={machine learning, multi-task learning, task affinity},
pdfproducer={Latex with hyperref},
pdfcreator={pdflatex}
]{hyperref}

\usepackage{cleveref}
\DeclareMathOperator*{\rank}{rank}
\DeclareMathOperator*{\argmax}{arg\,max}

\begin{document}

\maketitle

\begin{abstract}
While the promises of Multi-Task Learning (MTL) are attractive, characterizing the conditions of its success is still an open problem in Deep Learning. Some tasks may benefit from being learned together while others may be detrimental to one another. From a task perspective, grouping cooperative tasks while separating competing tasks is paramount to reap the benefits of MTL, \ie{} reducing training and inference costs. Therefore, estimating task affinity for joint learning is a key endeavor. Recent work suggests that the training conditions themselves have a significant impact on the outcomes of MTL. Yet, the literature is lacking of a benchmark to assess the effectiveness of tasks affinity estimation techniques and their relation with actual MTL performance. In this paper, we take a first step in recovering this gap by \textbf{(i) defining a set of affinity scores} by both revisiting contributions from previous literature as well presenting new ones and \textbf{(ii) benchmarking them} on the Taskonomy dataset. Our empirical campaign reveals how, even in a small-scale scenario, task affinity scoring does not correlate well with actual MTL performance. Yet, some metrics can be more indicative than others.
\end{abstract}

\section{Introduction}

For more than two decades since its inception \cite{caruana1997multitask}, Multi-Task Learning (MTL) has been extensively studied by the Deep Learning community. For practitioners interested in the best strategy to learn a collection of tasks, the promises of MTL are numerous and attractive. First, learning to solve several tasks simultaneously can be more cost-efficient from a model development and deployment perspective. Second, if the tasks learned together cooperate, MTL can even outperform its Single-Task Learning (STL) counterpart for the same computational cost \cite{standley2020tasks}.

However, MTL potential advantages are tempered by the difficulty of estimating \emph{task affinity}, \ie{} identify tasks benefiting from joint learning, without testing all combinations of tasks. 
This calls for \emph{task affinity scores} -- to quantify a priori and at a cheap computational cost the potential benefit of learning tasks together. 
The quest for the perfect affinity score is further exacerbated by MTL performance's strong dependency on the learning context, \ie{} the data and models used for training.
For instance, tasks cooperating in one learning context can result in competition
when using slightly different data or models~\cite{standley2020tasks}. 

Recent works \cite{fifty2021efficiently, standley2020tasks} have integrated this context-dependency when designing task grouping strategies. While these approaches avoid a complete search across all task combinations, they still require training and comparing some MTL models %in the end 
for the final network selection. Furthermore, those studies show that even in a small-scale scenario, MTL performance cannot be accurately predicted without actually performing MTL.

Despite providing assessment of task affinity, previous literature lacks of a broader comparison of the associated scores. In this work, we take a first step in recovering this gap by \textbf{presenting an empirical comparison of several task affinity scoring techniques}.
Some of these scores are inspired by previous literature ranging from Transfer Learning to Multi-Task Learning: \emph{taxonomical distance} \cite{zamir2018taskonomy}, \emph{input attribution similarity} \cite{input_attr_aff_metric_ref}, \emph{representation similarity analysis} \cite{TL_RSA}, \emph{gradient similarity} \cite{grads_clashes} and \emph{gradient transference} \cite{fifty2021efficiently}. We benchmark an additional affinity score which is an original proposal: \emph{label injection}. We evaluate all of them on the public Taskonomy dataset \cite{zamir2018taskonomy} which is a well-known large benchmark spanning several Computer Vision tasks. 
Note that our objective is not to present a novel state-of-the-art MTL architecture but rather an objective benchmark of task affinity estimation techniques. 
More specifically we aim to understand if task affinity scores can (i) be used as proxy for true MTL performance and (ii) suggest the best partner task to improve the performance of a target task.
These scores and their discussion aim at helping practitioners gauge the benefit of MTL for their own set of tasks.
In \cref{relatedwork}, we review the state of the art on MTL affinity characterization. In \cref{methodology}, we present the affinity scores selected for benchmarking and detail our evaluation protocol. We present our results in \cref{evaluation} and discuss the advantages and limitations of these scores in \cref{future}. \Cref{conclusion} concludes the paper.

\section{Background and related work} \label{relatedwork}
In this section, we first review relevant work on MTL and task grouping, briefly present the Taskonomy dataset, and finally introduce task affinity characterization. 

\paragraph{Multi-Task Learning}
The promises of MTL are based on the assumption that cooperative tasks benefit from inductive transfer during joint learning. 
By being learned together, tasks are encouraged to share, at least partially, common representations, \eg{} the extracted feature vector at the model's bottleneck, depending on the model architecture.
The intuition is that some tasks might exhibit compatible goals and help one another during training through synergies, \ie{} \emph{positive transfer}. However, tasks interference can still degrade performance if their respective updates become unaligned or contradictory during simultaneous learning \ie{} \emph{negative transfer} through competition. 

To mitigate these effects, two complementary lines of research both aim at reducing task interference and increasing task synergies. 
The first direction focuses on \emph{model design}, hence crafting the model such that it is adapted to learn a certain set of tasks. In this case, the task set is fixed while the model is adapted to fit all the tasks under consideration. Through hard parameter sharing, task weights can be adapted during training in order to balance their impact on the combined loss \cite{leang2020_dynamic_task_weighting_strats, pascal2021aaai}. Alternative approaches focus on tuning gradients to mitigate task interference during MTL training \cite{chen2018gradnorm, pcgrad, uncertaintyweights}. In soft parameter sharing instead, parameters are segregated by task and the model is guided, during MTL learning, to only share information when it is beneficial \cite{sun2020adashare_what_to_share, misra2016_cross_stitch}. 
The second research direction is more recent and focuses on \emph{task grouping strategies} by identifying cooperative tasks that can be grouped to be profitably learned together.
In this case, the model design is fixed while the task set is adapted \ie{} split into potentially overlapping subsets.
Recent works from \cite{fifty2021efficiently} and \cite{standley2020tasks} show promising results as they succeed in stimulating positive transfer by combining only tasks that are beneficial to one another. While these two research directions are complementary, our work is more in the scope of the latter as we benchmark affinity scores that should indicate if tasks benefit one another when learned together.

\paragraph{Task grouping}  

\begin{figure}[t]

\includegraphics[width=\columnwidth]{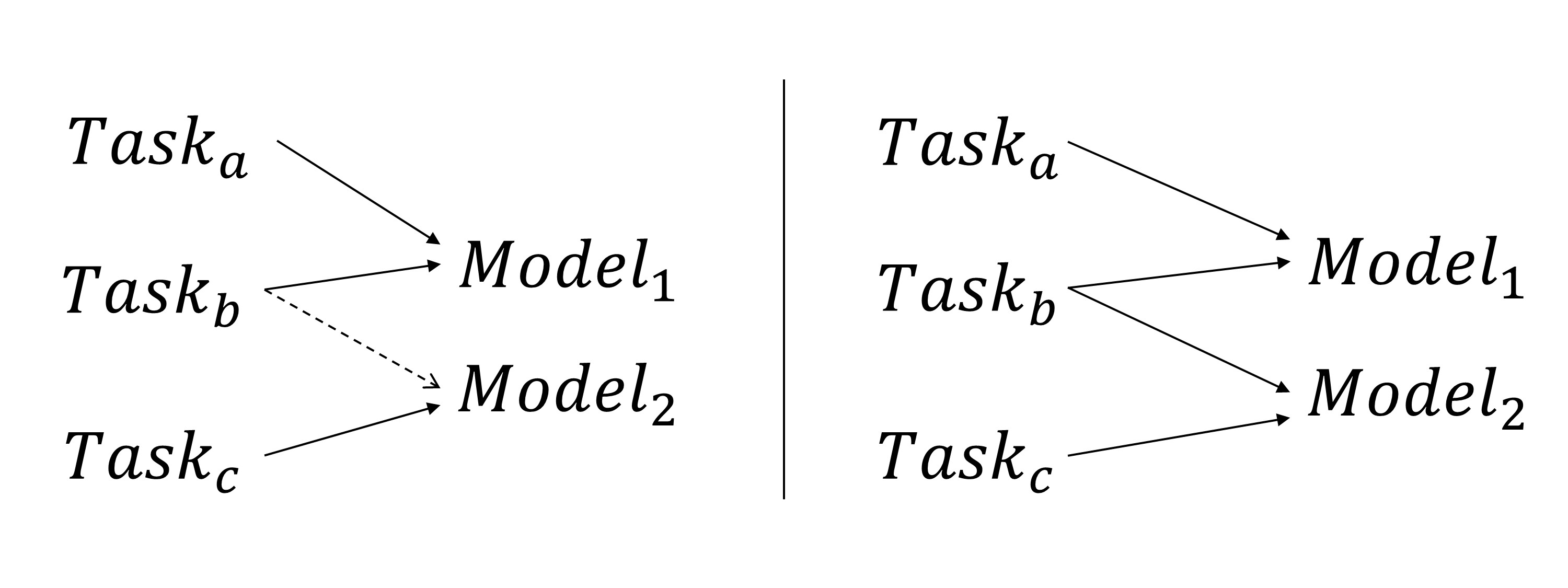}
\caption{(left) A \textit{valid} task grouping. Model$_1$ is assigned both Task$_a$ and Task$_b$ for inference. Model$_2$ is assigned Task$_c$ for inference and it uses Task$_b$ as a cooperative task only during training. (right) An \textit{invalid} task grouping. Model$_1$ and Model$_2$ are both assigned to solve Task$_b$ at inference time.}
\label{fig_tasks_groupings}
\end{figure}

Task grouping strategies aim at assigning tasks to models (that can be STL or MTL) in order to maximize the total performance of all the tasks under consideration, given a computational budget. More formally, let us consider the following: 
\begin{itemize}
    \item a set of $n$ tasks $\mathcal{T} = \{t_1, t_2, ..., t_n\}$ that need to be solved;
    \item a total computational budget of $\beta$ Multiply-Add operations;
    \item a set of $k \leq n$ models $\mathcal{M} = \{m_1, m_2, ..., m_k\}$, each one associated with its respective amount of Multiply-Adds operations $\mathcal{C} = \{c_1, c_2, ..., c_k\}$.
\end{itemize}
Task grouping aims at constructing $\mathcal{M}$ such that for all $t_i \in \mathcal{T}$ there exists exactly one model $m_j \in \mathcal{M}$ assigned to solve task $t_i$ at inference time, while respecting the computational budget $\sum_{j=1}^{k} c_j \le \beta$. Thus, any model from $\mathcal{M}$ can learn an arbitrary number of tasks as long as each task is assigned to one and only one model at inference time. 
To illustrate this point, we present valid and invalid task groupings in \cref{fig_tasks_groupings}.
The final objective of task grouping is to maximize the aggregated test performance: 
\begin{equation}
P = \sum_{t_i \in \mathcal{T}}^{}P(t_i|\mathcal{M}),
\end{equation}
where $P(t_i|\mathcal{M})$ denotes the performance\footnote{using a task-specific metric such as Intersection over Union for semantic segmentation or minimizing the model loss} of task $t_i$  using its assigned model from $\mathcal{M}$. It is worth mentioning that the task grouping problem differs from simple model selection as $(i)$ the objective (aggregated performance) and constraints (total cost) concern all tasks and models simultaneously and $(ii)$ any model can learn an arbitrary number of tasks.

The optimal task grouping is typically obtained by testing all task combinations within the computational budget. Therefore, to be as efficient as possible, grouping strategies rely on \emph{task affinity estimates} that guide the search of a solution in the task groups space. This approach might only identify sub-optimal task groups but it has a much a lower cost than an exhaustive search.
Sophisticated task grouping strategies are studied in \cite{fifty2021efficiently} in terms of performance and runtime. Such strategies include Higher-Order Approximation from \cite{standley2020tasks}, gradients cosine similarity maximization and task transference approximation from \cite{fifty2021efficiently}. Our work complements this benchmark of grouping strategies as we are interested in assessing the strengths and weaknesses of the \emph{underlying} affinity scores.
This includes an evaluation of the predictive quality of such scores. Indeed, the perfect scoring technique should not only identify the best partner tasks, but also be a proxy of the true MTL performance. 
Overall, we aim for a broader view of affinity scoring qualities with respect to what provided in the literature. 

\begin{figure}[t]
        \raisebox{0.5cm}{\includegraphics[width=\linewidth]{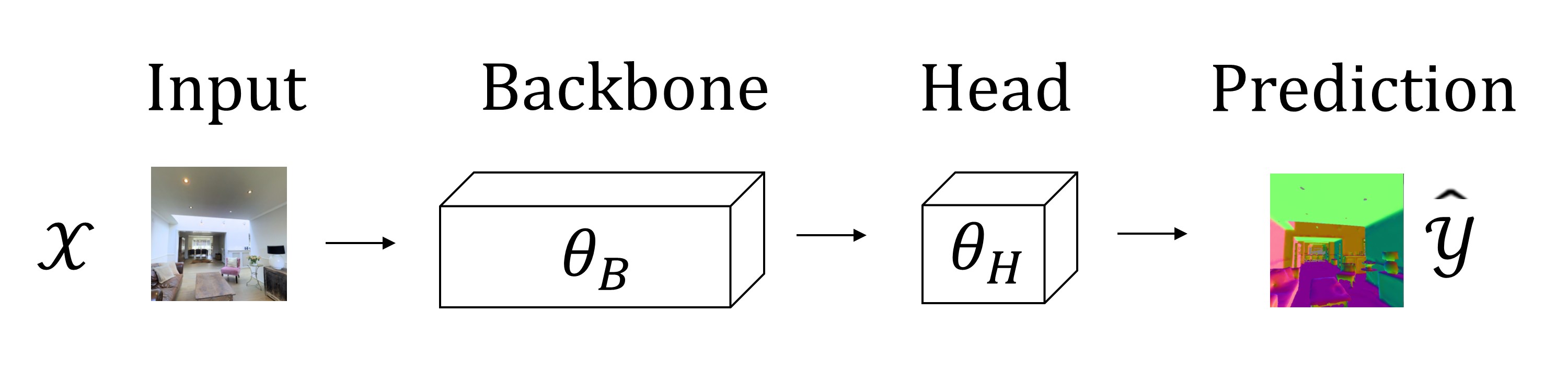}}
         \caption{STL model schematic architecture. $\theta_B$ denotes the backbone weights and $\theta_H$ the head weights.}
         \label{fig_stl}
\end{figure}

\paragraph{Taskonomy -- the reference framework}
From a Transfer Learning perspective, Taskonomy \cite{zamir2018taskonomy} has been a successful attempt at clarifying transfer synergies between visual tasks. From an MTL perspective, \cite{standley2020tasks} performs a broad empirical campaign on the same dataset to identify which visual tasks should be trained together with MTL. In particular, they evaluate if learning a target task with a partner task could outperform learning the target task alone. Thus, this framework quantifies task affinity as the performance gained on a task learned in MTL versus STL. First, the authors show that cooperation between tasks is not symmetrical, as one task may benefit from another but not necessarily the opposite. Second, by comparing MTL performance gains for the same pairs of tasks but learned in various settings \ie{} different dataset size or different MTL model capacity, they unveil the impact of the training context itself on task cooperation.
Based on this framework, \cite{fifty2021efficiently} monitors the evolution of task affinities during MTL training. 
Their experimentation on the CelebA dataset \cite{liu2018large} suggests that task cooperation evolves throughout training. 
Furthermore they also show that hyper-parameters such as the learning rate or the batch size can also affect cooperation.

Those works provide an in-depth view of relevant MTL training dynamics. Our work complements these findings with an in-breath view across several affinity scoring techniques that integrate, at varying degrees, data, model and hyper-parameters dependencies.

\paragraph{Task affinity}
We group the methods aiming at quantifying task affinity for MTL under the term ``affinity scores'' for short and break them down into three main categories depending on their requirements for computation.

\vspace{5pt}
\noindent \emph{Model-agnostic} affinity scores are computed using solely the data at hand. This may be accomplished using nomenclatures or taxonomies to loosely relate tasks. For example, Object classification and Semantic segmentation are both considered to be semantic tasks while Depth estimation is a 3-D task \cite{zamir2018taskonomy}. This can also be more sophisticated and make use of information theory to quantify how dependent two tasks are, \eg{} using labels entropy as in \cite{label_entropy_aff_metric_ref}.

\vspace{5pt}
\noindent\emph{STL-based} affinity scores make use of STL models and compare them to estimate affinity between tasks. Common approaches include comparing the STL models latent representations using \eg{} the Representation Similarity Analysis \cite{TL_RSA}. Another option is to compare the STL models attribution maps assuming cooperative tasks use the same features \cite{caruana1997multitask}. Also, drawing from Meta-Learning, \cite{task2vec} estimates affinity as the distance between tasks in an embedded space that encodes task complexity.

\vspace{5pt}
\noindent\emph{MTL-based} approaches estimate task affinity during the training of surrogate MTL model(s). 
Such computations need to be more efficient than testing all tasks combinations, otherwise it would defeat its very purpose of efficiently quantifying task affinity. \cite{fifty2021efficiently} proposes an affinity extraction method by simulating the effects that task-specific updates of the model parameters would have on other tasks. \cite{standley2020tasks} extends pairwise MTL performance gain to higher-order task combinations \ie{} groups of three or more tasks. Also, both \cite{fifty2021efficiently} and \cite{grads_clashes} propose to compute the cosine similarity between task-specific gradient updates as a way to estimate task affinity during MTL training.

\begin{figure}[t]
        \includegraphics[width=\linewidth]{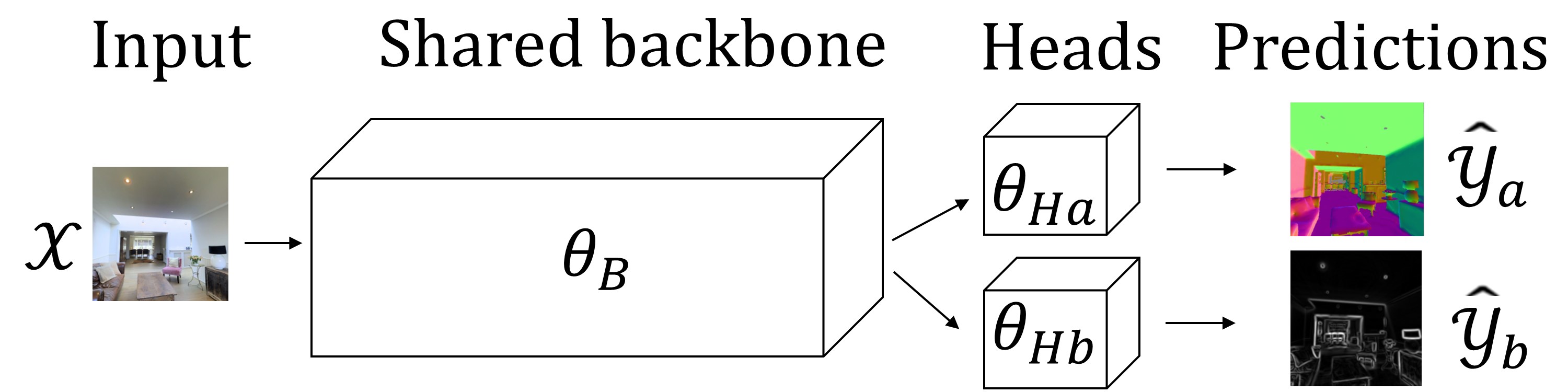}
         \caption{MTL model schematic architecture for two tasks $t_1=a$ and $t_2=b$. $\theta_B$ denotes the common backbone weights. $\theta_{Ha}$ and $\theta_{Hb}$ denote the separate heads weights.}
         \label{fig_mtl}
\end{figure}

\section{Methodology} \label{methodology}
Based on the assumption that grouping cooperative tasks together is a key success factor of MTL, we are interested in quantifying task affinity through several scores. In this section, we motivate the affinity scores selected and we detail the evaluation protocol implemented to benchmark them.

\subsection{Affinity scores} 

To simplify reasoning on task cooperation and competition, we restrict ourselves to \emph{pairwise} task affinity estimation, \ie{} affinity scores for 2-task MTL. We depict typical STL and pairwise-MTL architectures in Figures \ref{fig_stl} and \ref{fig_mtl} respectively. Considering two tasks $t_1=a$ and $t_2=b$ and a batch of examples $\mathcal{X}$, we denote: 
\begin{itemize}
    \item their resp. loss functions $\mathcal{L}_a$ and $\mathcal{L}_b$
    \item their resp. STL models $STL_a$ and $STL_b$ with losses 
    % $\mathcal{L}_{STL_a}$ and $\mathcal{L}_{STL_b}$
    \begin{itemize}
     \item $\mathcal{L}_{STL_a}=\mathcal{L}_a(\mathcal{X}, STL_a)$
     \item$\mathcal{L}_{STL_b}=\mathcal{L}_b(\mathcal{X}, STL_b)$
    \end{itemize}
    \item their joint MTL model $MTL_{(a,b)}$ with loss 
    \begin{itemize}
        \item
        $\mathcal{L}_{MTL_{(a,b)}}=\mathcal{L}_a(\mathcal{X}, MTL_{(a,b)}) + \mathcal{L}_b(\mathcal{X}, MTL_{(a,b)})$
    % = \mathcal{L}_{STL_a} + \mathcal{L}_{STL_b}$
    \end{itemize}
\end{itemize}
~\\
We consider six task affinity scores that we further describe in the remainder of this section. Their detailed computations are available in the supplementary material. Some scores are symmetric, \ie{} assessing how much two tasks $a$ and $b$ help each other regardless of direction; others instead are asymmetric, \ie{} assessing how much a target task $a$ benefits from being learned with a partner task~$b$. For each metric we report its category (model-agnostic, STL-based or MTL-based) and contribution (borrowed from literature, revisited from literature or novel).
~\\
\textbf{Taxonomical distance (TD)} \textit{Model-agnostic} -- \emph{borrowed}: A natural way of assessing affinity between tasks from a human perspective is to organize them through a hierarchical taxonomy. For example, classification datasets such as \cite{inaturalist_dataset} or \cite{cub_dataset} provide hierarchical class granularity that can be used to group similar tasks together as in \cite{task2vec}. In our case, we used the tasks similarity tree from \cite{zamir2018taskonomy}. This symmetric affinity score is computed as the distance between $a$ and $b$ in the tree.

~\\
\textbf{Input attribution similarity (IAS)} \textit{STL-based} -- \emph{revisited}: \cite{caruana1997multitask} defines related tasks as tasks that use the same features. Following this definition we assess how tasks relate to one another in terms of input attribution similarity using $Input X Gradient$ \cite{inputXgradient} to compute attribution maps for $STL_a$ and $STL_b$. The affinity score is then obtained via the cosine similarity of the attribution maps~\cite{input_attr_aff_metric_ref}. Therefore this score is symmetric.

~\\
\textbf{Representation similarity analysis (RSA)} \textit{STL-based} -- \emph{revisited}: RSA, a well-known method in the computational neuro-sciences community~\cite{TL_RSA}, relies on the assumption that, if tasks are similar, they learn similar representations, \ie{} a given input should be projected in similar locations in the latent space. Referring to \cref{fig_stl}, this score compares the latent representations structures between the respective backbones $\theta_{B}$ of $STL_a$ and of $STL_b$. In a nutshell, RSA uses the Spearman correlation of Representation Dissimilarity Matrices. This is a symmetric score.

~\\
\textbf{Label injection (LI)} \textit{STL-based} -- \emph{novel}: Another way to estimate task affinity is to measure the performance gained from adding the target label of another task to the input. For example, a task $a$ targeting the classification of handwritten digits could be paired with a task $b$ targeting the prediction of even and odd digits. Since the two tasks are (clearly) related, ``injecting'' the label of task $b$, \ie{} providing it as complementary input when training task $a$, could lead to performance increase for task $a$. The performance of label injection can be considered as a proxy of task affinity. This affinity score is asymmetric. It is computed as the performance gain between the standard STL model for task $a$ and the $b$-injected STL model for $a$ denoted by $STL_{a\leftarrow b}$, \ie{} 
\begin{equation}
    \frac{\mathcal{L}_{STL_a} - \mathcal{L}_{STL_{a\leftarrow b}}}{\mathcal{L}_{STL_{a\leftarrow b}}},
\end{equation}
using the test losses from the fully trained models.

~\\
\textbf{Gradient similarity (GS)} \textit{MTL-based} -- \emph{borrowed}: This task affinity score relies on the assumption that cooperative tasks yield similar \ie{} non-contradictory, weights updates to the model backbone during MTL training. This score, which we borrow from \cite{fifty2021efficiently, grads_clashes}, is symmetric. It is computed as the cosine similarity between gradients from each task loss with respect to the MTL model common backbone weights. Using the notation from \cref{fig_mtl}, we compute:
\begin{equation}
        S_{cos}\left(\frac{\partial{\mathcal{L}_a(\mathcal{X},\theta_B, \theta_{Ha})}}{\partial{\theta_{B}}}, \frac{\partial{\mathcal{L}_b(\mathcal{X},\theta_B, \theta_{Hb})}}{\partial{\theta_{B}}}\right),
\end{equation}
at each epoch, and average these cosine similarities across all training epochs.

~\\
\textbf{Gradient transference (GT)} \textit{MTL-based} -- \emph{borrowed}: During MTL training, by simulating task-specific updates to the common backbone, one can estimate how it would impact the other task's performance. This corresponds to the losses look-ahead ratio defined in \cite{fifty2021efficiently}. This asymmetric score is computed comparing the loss of task $a$ after updating the common backbone according to $b$, and the loss of task $a$ before this simulated update. Referring to the notation from \cref{fig_mtl}, we denote the $b$-specific update of the common backbone by $\theta_{B|b}$. Thus, we compute:
\begin{equation}
        \frac{\mathcal{L}_a(\mathcal{X},\theta_{B|b}, \theta_{Ha})}{\mathcal{L}_a(\mathcal{X},\theta_{B}, \theta_{Ha})},
\end{equation}
at each epoch, and average these ratios throughout training.

\subsection{Evaluation}
\label{methodology_evaluation}

We evaluate these affinity scores against the true MTL performance. Moreover, we evaluate the scores across three levels by progressively relaxing the constraint of the analysis.

~\\
\textbf{True performance: MTL gain.} 
As in \cite{standley2020tasks}, we quantify MTL success as the relative gain between STL and MTL performance in terms of test loss. MTL gain for a target task $a$ when using a partner task $b$ is defined as:
\begin{equation}
    \mathcal{G}(a|b) = \frac{\mathcal{L}_{STL_a} - \mathcal{L}_{MTL_{(\bm{\bar{a}},b)}}}{\mathcal{L}_{MTL_{(\bm{\bar{a}},b)}}},
\end{equation}
where $\mathcal{L}_{STL_a}$ is the test loss for task $a$ in a STL configuration, and $\mathcal{L}_{MTL_{(\bm{\bar{a}},b)}}$ is the test loss for task $a$ in a MTL configuration using tasks $a$ and $b$ for joint learning. Note that the contribution of task $b$ to the MTL loss is not considered when computing the gain, yet is considered during training. We perform an exhaustive search through all possible pairs of tasks to compute the ``ground truth'' affinities. These serve as baseline against which each affinity score is evaluated.

~\\
\textbf{Level 1: predictive power.}
As previously stated, an ideal affinity score should be a proxy of the actual MTL gain: higher/lower score should imply large/small benefit from joint training.
This is a stringent requirement, yet easy to quantify by mean of Pearson's correlation. Specifically, for each target task $a$ and affinity scoring technique, we compute the correlation between the MTL gain across all partner tasks (the true performance) and the affinity score across the same partners (the proxy of the performance). It follows that affinity scoring techniques with correlation values close to $-1$ (perfect negative correlation) and $+1$ (perfect positive correlation) have strong predictive power; correlation values close to zero imply no predictive power.

~\\
\textbf{Level 2: partners ranking.} 
To relax the previous requirement, we define acceptable an affinity score capable to successfully rank potential partner tasks by decreasing order of MTL gain. More formally, for a target task $a$, and a set of partner tasks $\mathcal{P}$, we want an affinity score $\delta$ such that: 
\begin{equation}
    \forall{\, t_i} \in \mathcal{P}, \, \rank(\,\delta(a, t_i)\,) = \rank(\,\mathcal{G}(a|t_i)\,),
\end{equation}
To evaluate the agreement between the ranking given by the affinity score and the actual ranking by MTL gain obtained by exhaustive search, we use Kendall's correlation coefficient \cite{kendall1948rank} that ranges from $-1$ (opposite rankings) to $+1$ (same rankings).

~\\
\textbf{Level 3: best partner identification.} 
In case only pairs of tasks are considered for MTL, one is essentially interested in finding the best partner. This means that we can further relax the previous constraint and for a target task $a$, we want an affinity score $\delta$ such that:
\begin{equation}
    \argmax_{t_i\in\mathcal{P}}{\delta(a,t_i)} = \argmax_{t_i\in\mathcal{P}}{\mathcal{G}(a|t_i)},
\end{equation}
To evaluate this, we report the MTL gain obtained when choosing the top partner according to the affinity score and compare it with the maximum MTL gain obtained when choosing the actual best partner. 

\section{Results} \label{evaluation}

In this section we first detail the data and models used to evaluate the proposed affinity scores. Then, we present the results of our empirical campaign along the three levels of evaluation previously defined.

\subsection{Experimental protocol}

~\\
\textbf{Dataset.} 
In this work, we select a portion of the Taskonomy medium-size split. This constitutes a representative dataset of Computer Vision tasks, composed of labeled indoor scenes from 73 buildings whose list is available in the supplementary material. The whole dataset amounts to 726,149 input images which represent approximately 1.2 TB including the various labels. We select the same five tasks as \cite{standley2020tasks, fifty2021efficiently} to conduct our experiments, namely: 
\begin{itemize}
    \item Semantic segmentation (\emph{SemSeg})
    \item 2D SURF keypoints identification (\emph{Keypts})
    \item Edges texture detection (\emph{Edges})
    \item Depth Z-Buffer estimation (\emph{Depth})
    \item Surface normals estimation (\emph{Normal})
\end{itemize}
A detailed description of the tasks can be found in the supplementary material from \cite{zamir2018taskonomy}.

~\\
\textbf{Models definition.} 
We build on the work of \cite{standley2020tasks} to train five STL models for the five aforementioned tasks and ten pairwise MTL models. Models are variants of the Xception architecture \cite{chollet2017xception}, composed of a backbone that learns a latent representation of the input and a head. In the case of the STL models, the backbone output is forwarded to a single head that produces the final prediction, cf. \cref{fig_stl}. In the case of the pairwise MTL models, the shared backbone output is forwarded to two disjoint heads, one for each task under consideration by the MTL model, cf. \cref{fig_mtl}. In this work, as well as in \cite{standley2020tasks, fifty2021efficiently}, we only consider hard parameter sharing for the MTL backbone. While this approach simplifies reasoning about shared representations and weights updates, it does not incorporate task interference mitigation strategies. 

In terms of model capacity, we replicate the Xception17 models design from prior work in \cite{standley2020tasks}, allowing each STL model only half of the capacity \ie{} number of Multiply-Add operations, of a pairwise MTL model. This constraint is implemented by reducing the number of channels in the CNN blocks composing the backbone. Therefore, STL and MTL models use the same architecture but with varying capacity. Each model is trained for 50 epochs with a decreasing learning rate, selecting the best-performing epoch on the validation set as final model. Finally, hyper-parameters are set to default values from \cite{standley2020tasks} with no further tuning.

\subsection{Experimental results}

In the following we report our evaluation based on the methodology described in \cref{methodology_evaluation}. The detailed values of each affinity score are instead reported in the supplementary material. 

\begin{table}[t]
\small
\centering
\setlength{\tabcolsep}{3pt}
\newcolumntype{Y}{>{\centering\arraybackslash}X}
    \begin{tabularx}{\linewidth}{
    Y|
    Y
    Y
    Y
    Y
    Y|
    Y
    }
    
    \multicolumn{1}{c|}{} &
    \multicolumn{5}{c|}{\textbf{MTL gain on}} &
    \multicolumn{1}{c}{}  \\
    \multicolumn{1}{c|}{\makecell{Trained\\with}} &
    \multicolumn{1}{c}{\textbf{SemSeg}} &
    \multicolumn{1}{c}{\textbf{Keypts}} &
    \multicolumn{1}{c}{\textbf{Edges}} &
    \multicolumn{1}{c}{\textbf{Depth}} & 
    \multicolumn{1}{c|}{\textbf{Normal}} &
    \multicolumn{1}{c}{\textbf{Avg.}} \\
    
    \hline
    
    SemSeg & - & -11.81 & -10.22 & -0.55 & +0.95 & -5.41 \\ 
    Keypts &-6.70& - & -8.67 & -9.87 & -13.88 & -9.78 \\ 
    Edges & -22.01 & +1.26 & - & -8.24 &  +2.18 & -6.70 \\ 
    Depth & +18.02 & -3.81 & +16.69 & - &  -6.37 & +6.13 \\ 
    Normal & +50.24 & +29.56 & +78.05 & -0.45 & - & +39.35 \\ 
    
    \end{tabularx}
    \caption{\textit{True performance: MTL gain.} Ground-truth MTL gain for each target task (column) and each partner task (row) \eg{} the task Edges performs 78.05\% better than learned alone in STL when trained with Normal as partner.}
    \label{setting3bis_gt_affinities}
\end{table}

~\\
\textbf{MTL gain.} 
In \Cref{setting3bis_gt_affinities}, we report the ground truth MTL gain for each pair of tasks. We reiterate that these results serve as reference for evaluating the affinity scores. Furthermore, recall that MTL gains are tightly related to the specific training conditions of our experiment \ie{} the data, models and hyper-parameters used, and they may vary if computed in another setting. From this table, we note that some tasks are more helpful than others. For example, \emph{Normal} is a helpful partner task, but fails to be significantly assisted by any other task. Overall, we find MTL gains to be highly asymmetric. Nonetheless, almost all tasks would benefit from being learned with their best partner. This is in line with the findings of \cite{standley2020tasks}.

\begin{table}[ht!]
\small
\centering
\setlength{\tabcolsep}{3pt}
\newcolumntype{Y}{>{\centering\arraybackslash}X}
    \begin{tabularx}{\linewidth}{
    >{\hsize=.4\hsize}Y|
    >{\hsize=.23\hsize}Y|
    >{\hsize=.23\hsize}Y
    >{\hsize=.23\hsize}Y
    >{\hsize=.23\hsize}Y|
    >{\hsize=.23\hsize}Y
    >{\hsize=.23\hsize}Y
    }
    \multicolumn{1}{c|}{} &
    \multicolumn{1}{c|}{\makecell{Model\\agnostic}} &
    \multicolumn{3}{c|}{STL-based} &
    \multicolumn{2}{c}{MTL-based}  \\
    
    \textbf{\makecell{Task}} & 
    \textbf{TD} & 
    \textbf{IAS} & \textbf{RSA} & \textbf{LI} &
    \textbf{GS} & \textbf{GT} \\
     
     \hline
     
    SemSeg & 0.4
    & 0.99
    & 0.81 
    & 0.99
    & 0.79
    & 0.76 
    \\ 
     
    Keypts & -0.03
    & -0.06
    & -0.37
    & 0.95
    & 0.22
    & -0.08 
    \\ 
    
    Edges & -0.34 
    & -0.44
    & -0.68 
    & 0.90
    & -0.37
    & -0.66 
    \\
    
    Depth & 0.90
    & 0.98
    & 0.96
    & 0.64 
    & 0.69
    & 0.97
    \\ 
    
    Normal & 0.60 
    & 0.38
    & 0.20 
    & -0.11 
    & -0.19
    & 0.40
    \\ 
    
    \hline
    
    All-at-once & 0.08 
    & 0.08 
    & -0.15
    & \textbf{0.47}
    & -0.08
    & -0.02 
    \\ 

    \end{tabularx}
    \caption{\textit{Level 1: predictive power.} Affinity scores correlation with MTL gain. 
    \eg{} using Label injection (LI) to estimate affinities for the target task SemSeg, its output strongly correlates with the actual MTL gains (Pearson corr. $= 0.99$).}
  \label{evaluation_corr_gt}
\end{table}

~\\
\textbf{Predictive power.} 
\Cref{evaluation_corr_gt} %
shows the Pearson correlation between the MTL gain and each individual affinity score. Each row considers a separate target task, while the last row labeled as \emph{all-at-once} reports the correlation computed using all pairs of all target tasks together.

Starting from such an aggregate scenario, we can see that no scoring technique strongly correlates with the MTL gains. Only \emph{Label injection} moderately correlates with MTL gain across all tasks pairs (Pearson corr. $= 0.47$). This invalidates the predictability property desired for an ideal affinity score. Interestingly, when considering a single target task at a time, some affinity scores successfully predict MTL performance. For example, \emph{Depth}'s MTL gains can be predicted using \emph{Input attribution similarity} (corr. $= 0.98$). Yet, no scoring provides a stable correlation across \emph{all} tasks pairs.

~\\
\textbf{Partners ranking.} 
In \Cref{affinities_ranking}, we evaluate each affinity scoring technique in terms of its ability to correctly rank potential partner tasks according to the MTL gains they provide. For a given target task, we compare the rank obtained from the affinity score with the rank obtained from the MTL gains by mean of the Kendall rank correlation. As in the predictive power evaluation table, each row reports on the correlation for each target task separately while in this case the last line summarizes the overall performance using the average of the rank correlations across target tasks.

Starting from the aggregate view, we observe that no score-based ranking correlates strongly with true ranking. Only \emph{Label injection} and \emph{Gradients similarity} show a moderate and positive correlation (average Kendall corr. $=0.47$ and $0.4$ resp.). Differently from before, when considering specific targets tasks, the correlation does not necessarily improve. For instance, \emph{Keypts} and \emph{Normal} STL-based scores completely fail, yet MTL-based scores are not necessarily better. Still considering \emph{Keypts} target task, notice how \emph{Label injection} shows significantly higher Pearson correlation, while the Kendall correlation shows that half of the partner tasks are wrongly ranked according to the affinity score.

\begin{table}[h]   
\small
\centering
\setlength{\tabcolsep}{3pt}
\newcolumntype{Y}{>{\centering\arraybackslash}X}
    \begin{tabularx}{\linewidth}{
    Y|
    Y|
    Y
    Y
    Y|
    Y
    Y
    }
    \multicolumn{1}{c|}{} &
    \multicolumn{1}{c|}{\makecell{Model\\agnostic}} &
    \multicolumn{3}{c|}{STL-based} &
    \multicolumn{2}{c}{MTL-based}  \\
    
    \textbf{\makecell{Task}} & 
    \textbf{TD} & 
    \textbf{IAS} & \textbf{RSA} & \textbf{LI} &
    \textbf{GS} & \textbf{GT} \\
     
     \hline
     
    SemSeg & 0.0 
    & 1.0
    & 0.33 
    & 1.0
    & 0.67
    & 0.67 
    \\ 
     
    Keypts & 0.0 
    & 0.0
    & 0.0 
    & 0.0 
    & 0.67
    & 0.33 
    \\ 
    
    Edges & -0.33 
    & -0.33
    & -0.33 
    & 0.67
    & 0.0
    & -0.33 
    \\ 
    
    Depth & 1.0
    & 0.67 
    & 1.0
    & 0.67 
    & 1.0
    & 0.67 
    \\ 
    
    Normal & 0.33 
    & 0.0
    & 0.0 
    & 0.0 
    & -0.33
    & 0.0 
    \\ 
    
    \hline
    
    Average & 0.2 
    & 0.27 
    & 0.2
    & \textbf{0.47}
    & 0.4
    & 0.27 
    \\ 
    
    \end{tabularx}
    \caption{\textit{Level 2: partners ranking.} Comparison of partner tasks ranking by affinity score versus by MTL gain. 
    \eg{} Label injection (LI) perfectly ranks partners for the target task SemSeg (Kendall corr.$= 1$).}
    \label{affinities_ranking}
\end{table}

\begin{table*}[t!]   
\small
\centering
\setlength{\tabcolsep}{1.5pt}
\newcolumntype{Y}{>{\centering\arraybackslash}X}
    \begin{tabularx}{\textwidth}{
    >{\Centering \hsize=.19\hsize}Y
    >{\Centering \hsize=.23\hsize}Y|
    >{\Centering \hsize=.30\hsize}Y|
    >{\Centering \hsize=.23\hsize}Y
    >{\Centering \hsize=.23\hsize}Y
    >{\Centering \hsize=.23\hsize}Y|
    >{\Centering \hsize=.23\hsize}Y
    >{\Centering \hsize=.23\hsize}Y
    }
    
    \multicolumn{2}{c|}{} &
    \multicolumn{1}{c|}{Model-agnostic} &
    \multicolumn{3}{c|}{STL-based} &
    \multicolumn{2}{c}{MTL-based}  \\
    
    \textbf{Task} & \textbf{Expected partner} &
    \textbf{TD} & 
    \textbf{IAS} & \textbf{RSA} & \textbf{LI} &
    \textbf{GS} & \textbf{GT} \\
     
     \hline
     
    \textbf{SemSeg} & Normal & Normal (0) & Normal (0) & Depth (-32.2) & Normal (0) & Depth (-32.2)& Depth (-32.2)\\ 
     
    \textbf{Keypts} & Normal & Edges (-28.3) & Edges (-28.3) & Edges (-28.3) & Normal (0) & Edges (-28.3) & Edges (-28.3) \\ 
    
    \textbf{Edges} & Normal & Keypts (-86.7) & Keypts (-86.7) & Keypts (-86.7) & Normal (0) & Keypts (-86.7) & Keypts (-86.7) \\ 
    
    \textbf{Depth} & Normal & Normal (0) & SemSeg (-0.1) & Normal (0) & Normal (0) & Normal (0) & SemSeg (-0.1) \\ 
    
    \textbf{Normal} & Edges & SemSeg/Depth (-4.9) & SemSeg (-1.2) & Depth (-8.6) & Depth (-8.6) & Depth (-8.6) & Depth (-8.6)\\ 
    
    \hfill\\
    
    \end{tabularx}
    \caption{\textit{Level 3: best partner identification}.
    Comparison of best partner selection by affinity score. In parenthesis, we report the difference of MTL gain between the actual best and the selected partner. 
    %Rounded at the first decimal. 
    \eg{} For the target task Keypts the actual best is Normal and all scores but Label injection (LI) select Edges leading to 1.26 - 29.56 = -28.3 decrease in performance gain.
    }
    \label{affinities_best_partner}
\end{table*}

~\\
\textbf{Best partner identification.} 
\Cref{affinities_best_partner} shows the top-1 partner according to each affinity scoring technique. This is to be compared with the maximum MTL gain that can be achieved using the actual best partner. \emph{Label Injection} correctly identifies the best partner for four out of five tasks. However, not a single affinity score is capable of correctly identifying \emph{Normal}'s best partner for MTL. Furthermore, \emph{Keypts} and \emph{Edges} seem to be particularly difficult tasks for best partner identification. All scores but \emph{Label injection} recommend choosing either one as best partner for the other, while the actual best choice is \emph{Normal} for both of them.

\section{Discussion and future work} \label{future}
A perfect affinity score should be both predictive of the actual MTL gain and cheap to compute. As prior work hints that the training conditions themselves impact MTL gain, it seems particularly tough to reconcile these properties as we also verify throughout our experimental campaign. In this paper, we benchmark various affinity scoring techniques that incorporate data, model and hyper-parameters dependencies at varying degrees: from model-agnostic scores that do not take these into account, through STL-based scores that try to include them, to MTL-based scores that are supposed to be the closest to the actual MTL learning conditions. Unfortunately, none of the selected scores, not even the MTL-based ones that are close to MTL training, can accurately predict MTL gain across all pairs of tasks. However, \emph{Label injection}, the original affinity score we introduce, appears useful for predicting the gains corresponding to potential partners given a target task. We also observe that, surprisingly, MTL-based scores are not necessarily better than STL-ones, \ie{} not even quantifying affinity during the actual MTL training seems sufficient to link affinity to performance.

From a cost perspective, except for \emph{Taxonomical distance}, all the scoring techniques we benchmark require some model training. We quantify the computational cost of an affinity score by the total amount of training it requires to estimate affinities across \emph{all} pairs of tasks. Let us consider $n$ tasks and a standard half-capacity STL model with its respective number of Multiply-Add operations denoted by $c_{s}$. Using this notation, we report the computational cost associated with each affinity score in Table \ref{costs}. While \emph{Input attribution similarity} and \emph{Representation similarity} only require one STL model per task\footnote{In some scenario, the STL models may be readily available, such that the costs associated with \emph{Input attribution similarity} and \emph{Representation similarity} can be amortized.}, \emph{Label injection} requires to train an additional STL-injected model for each ordered pair. Regarding MTL-based scores, both \emph{Gradient similarity} and \emph{Gradient transference}\footnote{We neglect the cost of the simulated task-specific update during \emph{Gradient transference} training.} require to train a full-capacity MTL model for each unordered pair of tasks. Finally, while \emph{Taxonomical distance} may appear cost-efficient, it has been established using a Transfer Learning-based taxonomy that itself requires STL models training, cf. \cite{zamir2018taskonomy}.

\begin{table}[!t]
\small
\centering
\setlength{\tabcolsep}{1.5pt}
\newcolumntype{Y}{>{\centering\arraybackslash}X}
    \begin{tabularx}{\linewidth}{
    >{\hsize=0.6\hsize}Y|
    >{\hsize=0.3\hsize}Y|
    >{\hsize=0.3\hsize}Y|
    >{\hsize=0.3\hsize}Y|
    >{\hsize=0.5\hsize}Y|
    >{\hsize=0.4\hsize}Y|
    >{\hsize=0.4\hsize} Y
    }
    \multicolumn{1}{c|}{} &
    \multicolumn{1}{c|}{\makecell{Model\\agnostic}} &
    \multicolumn{3}{c|}{STL-based} &
    \multicolumn{2}{c}{MTL-based} \\
    
    \textbf{Cost} &
    \textbf{TD} &
    \textbf{IAS} &
    \textbf{RSA} &
    \textbf{LI} & 
    \textbf{GS} &
    \textbf{GT} \\

    \hline
    
    \makecell{$\#$ of \\ Multiply \\ Adds} & 
    0 & 
    $n \cdot c_s$ & 
    $n \cdot c_s$ &
    $n \cdot c_s + 2{n \choose 2} \cdot c_s$ & 
    ${n \choose 2} \cdot 2 c_s$ & 
    ${n \choose 2} \cdot 2 c_s$ \\
    
    \end{tabularx}
    \caption{\textit{Affinity scores costs.} Comparison of training costs considering all pairs across $n$ tasks, where $c_{s}$ denotes the amount of Multiply-Add operations for a standard half-capacity STL model.}
    \label{costs}
\end{table}

From a practical perspective, \emph{Label injection} can correctly identify the best partner for most tasks, except for \emph{Normal}, for which none of the other scores succeeded. As in \cite{standley2020tasks}, we find that \emph{Normal} is different from the other tasks: it benefits others but it is better learned alone. We conjecture that the high complexity of this task makes it a good partner for sharing knowledge during joint learning, but prevents it from being helped by easier tasks. To further corroborate this hypothesis, task complexity need to be incorporated in the affinity scoring design. We believe that \textit{Task2Vec} from \cite{task2vec} is a first step towards this direction as it establishes a distance metric between tasks incorporating task difficulty from a Transfer Learning perspective. Unfortunately, \textit{Task2Vec} cannot be directly used in our context as it has only been defined for homogeneous tasks \ie{} from the same domain. Indeed, in \cite{task2vec}, the tasks are defined using coarse or fine-grained classification variations from the same hierarchy. We leave the exploration of this research direction as future work. 

While this empirical campaign provides a better understanding of the challenges to take up when designing task affinity scores, it is not conclusive given the high variability coming from data, models, and tasks used for MTL. In other words, while some results are encouraging, more research is required to make those mechanisms actionable for an actual model design and operation. In this direction, we identify some limitations that we intend to tackle in future work. First, this analysis is limited to five Computer Vision tasks. Some model-agnostic affinity scores such as \emph{Taxonomical distance} might not be trivially adapted to other task domains. Second, the affinity scores we defined can only estimate pairwise task affinity. While this is a reasonable starting point, various effects may be at play when learning more than two tasks simultaneously. \cite{aaai_disentangling} propose a new perspective on a sample-wise basis to quantify task transfer and interference separately. However, their metrics are defined for classification tasks only and their adaptation to heterogeneous tasks is an open question. Third, the MTL architecture we selected features hard parameter sharing and a static combined loss with equal weights. Although this design choice is consistent with prior work and facilitates reasoning on tasks cooperation, it does not take advantage of the recent advances in task interference mitigation techniques for MTL training \cite{chen2018gradnorm, pcgrad, uncertaintyweights}. Indeed, tasks may be affine but still interfere during joint learning if no mechanism is implemented to attenuate it, which is why MTL architecture design and task grouping strategies are complementary lines of research. 

\section{Conclusion} \label{conclusion}
Based on the assumption that identifying cooperative tasks to be learned together is a key success factor of MTL, we borrowed, adapted and designed various task affinity scores for this purpose. We benchmarked these scores for pairs of tasks on a public Computer Vision dataset to discuss their strengths and weaknesses. Although no score is perfectly predictive of MTL gain, some of them still hold value for practitioners, by being able to identify the best partner for a given target task. This empirical campaign offers a better understanding of the conditions that allow MTL to be superior to STL and sheds light on the challenges to be met when predicting it.
\bibliography{main}

\clearpage
\appendix
\begin{abstract}
The following elements are provided in the supplementary material:
\begin{enumerate}
    \item Affinity scores raw values
    \item Taskonomy buildings used
    \item Affinity scores computation
\end{enumerate}
\end{abstract}

\section{Affinity scores raw values} \label{appendix:aff_metrics_raw_results}
In Tables \ref{appendix:aff_metric_td} to \ref{appendix:aff_metric_gt}, we report the raw affinities estimations for all tasks, using each affinity scoring technique. Results are rounded at the second decimal.

~\\
\begin{table}[h!]    
\small
\centering
\setlength{\tabcolsep}{3pt}
\newcolumntype{Y}{>{\centering\arraybackslash}X}
    \begin{tabularx}{\linewidth}{
    Y|
    Y
    Y
    Y
    Y
    Y
    }
    
    \multicolumn{1}{c|}{} &
    \multicolumn{5}{c}{\textbf{Affinities estimations}} \\
    
    \multicolumn{1}{c|}{with} & 
    \multicolumn{1}{c}{\textbf{SemSeg}} & 
    \multicolumn{1}{c}{\textbf{Keypts}} & 
    \multicolumn{1}{c}{\textbf{Edges}} & 
    \multicolumn{1}{c}{\textbf{Depth}} & 
    \multicolumn{1}{c}{\textbf{Normal}} \\ 

    \hline
     
    SemSeg & - & -8 & -6 & -8 & -5 \\ 
     
    Keypts & -8 & - & -4 & -12 & -9  \\ 
    
    Edges & -6 & -4 & - & -10 & -7 \\ 
    
    Depth & -8 & -12 & -10 & - & -5  \\ 
    
    Normal & -5 & -9 & -7 & -5 & - \\ 
    
    \end{tabularx}
    \caption{\textit{Taxonomical distance (TD).} Distance between tasks in the similarity tree from \cite{zamir2018taskonomy}. Multiplied by $-1$ for consistency (\ie{} higher means more affinity).}
    \label{appendix:aff_metric_td}
\end{table}

\begin{table}[h!]    
\small
\centering
\setlength{\tabcolsep}{3pt}{}
\newcolumntype{Y}{>{\centering\arraybackslash}X}
    \begin{tabularx}{\linewidth}{
    Y|
    Y
    Y
    Y
    Y
    Y
    }
    
    \multicolumn{1}{c|}{} &
    \multicolumn{5}{c}{\textbf{Affinities estimations}}  \\
    
    \multicolumn{1}{c|}{with} &
    \multicolumn{1}{c}{\textbf{SemSeg}} & 
    \multicolumn{1}{c}{\textbf{Keypts}} & 
    \multicolumn{1}{c}{\textbf{Edges}} &
    \multicolumn{1}{c}{\textbf{Depth}} & 
    \multicolumn{1}{c}{\textbf{Normal}}  \\

    \hline
     
    SemSeg &   -   & 0.25 & 0.23 & 0.31 & 0.45 \\ 
     
    Keypts & 0.25 &   -  & 0.52 & 0.18 & 0.23  \\ 
    
    Edges  & 0.23 & 0.52 &   -   & 0.18 & 0.22 \\ 
    
    Depth  & 0.31 & 0.18 & 0.18 &  -   & 0.29  \\ 
    
    Normal & 0.45 & 0.23 & 0.22 & 0.29 &   -    \\ 
    
    \end{tabularx}
    \caption{\textit{Input attribution similarity (IAS).} Cosine similarity between STL models attribution maps. 
    }
    \label{appendix:aff_metric_ias}
\end{table}

\begin{table}[h!]   
\small
\centering
\setlength
\setlength{\tabcolsep}{3pt}
\newcolumntype{Y}{>{\centering\arraybackslash}X}
    \begin{tabularx}{\linewidth}{
    Y|
    Y
    Y
    Y
    Y
    Y
    }
    
    \multicolumn{1}{c|}{} &
    \multicolumn{5}{c}{\textbf{Affinities estimations}}  \\
    
    \multicolumn{1}{c|}{with} & 
    \multicolumn{1}{c}{\textbf{SemSeg}} & 
    \multicolumn{1}{c}{\textbf{Keypts}} & 
    \multicolumn{1}{c}{\textbf{Edges}} & 
    \multicolumn{1}{c}{\textbf{Depth}} & 
    \multicolumn{1}{c}{\textbf{Normal}}  \\ 

    \hline
     
    SemSeg &  -   & 0.33 & 0.37 & 0.46 & 0.46 \\ 
     
    Keypts & 0.33 &   -   & 0.66 & 0.04 & 0.05  \\ 
    
    Edges  & 0.37 & 0.66 &   -   & 0.12 & 0.13 \\ 
    
    Depth  & 0.46 & 0.04 & 0.12 &   -   & 0.69  \\ 
    
    Normal & 0.46 & 0.05 & 0.13 & 0.69 &   -    \\ 
    
    \end{tabularx}
    \caption{\textit{Representation similarity analysis (RSA).} Representation similarity analysis using the STL models backbones output. 
    }
    \label{appendix:aff_metric_rsa}
\end{table}

\begin{table}[h!]   
\small
\centering
\setlength{\tabcolsep}{3pt}
\newcolumntype{Y}{>{\centering\arraybackslash}X}
    \begin{tabularx}{\linewidth}{
    Y|
    Y
    Y
    Y
    Y
    Y
    }
    
    \multicolumn{1}{c|}{} &
    \multicolumn{5}{c}{\textbf{Affinities estimations}}  \\
    
    \multicolumn{1}{c|}{with} & 
    \multicolumn{1}{c}{\textbf{SemSeg}} & 
    \multicolumn{1}{c}{\textbf{Keypts}} & 
    \multicolumn{1}{c}{\textbf{Edges}} & 
    \multicolumn{1}{c}{\textbf{Depth}} & 
    \multicolumn{1}{c}{\textbf{Normal}} \\ 

    \hline
     
    SemSeg & -  & -2.93  & -3.50  & -3.07 & +3.70 \\ 
     
    Keypts & -8.47  &   -   & +4.97  & -8.31 & +1.42  \\ 
    
    Edges  & -15.93 & -4.20  &   -   & -9.58 & +2.42 \\ 
    
    Depth  & +4.04  & -3.34  & -1.26  &  -  & +20.29  \\ 
    
    Normal & +25.68 & +60.29 & +23.79 & +66.30 &  - \\ 
    
    \end{tabularx}
    \caption{\textit{Label injection (LI).} Performance gain (\%) when incorporating the label from the partner task in the STL model's input, relative to standard STL.}
    \label{appendix:aff_metric_li}
\end{table}

\begin{table}[h!]    
\small
\centering
\setlength{\tabcolsep}{3pt}
\newcolumntype{Y}{>{\centering\arraybackslash}X}
    \begin{tabularx}{\linewidth}{
    Y|
    Y
    Y
    Y
    Y
    Y
    }
    
    \multicolumn{1}{c|}{} &
    \multicolumn{5}{c}{\textbf{Affinities estimations}}  \\
    
    \multicolumn{1}{c|}{with} &
    \multicolumn{1}{c}{\textbf{SemSeg}} & 
    \multicolumn{1}{c}{\textbf{Keypts}} & 
    \multicolumn{1}{c}{\textbf{Edges}} & 
    \multicolumn{1}{c}{\textbf{Depth}} & 
    \multicolumn{1}{c}{\textbf{Normal}}  \\ 

    \hline
    
    SemSeg &  -  & 0.51    & 0.39  & 1.93 & 1.54 \\ 
     
    Keypts & 0.51  &  -   & 1.89  & 0.75 & 1.0  \\ 
    
    Edges  & 0.39  & 1.89   &   -    & 0.92 & 0.59 \\ 
    
    Depth  & 1.93  & 0.75   & 0.92  &    -   & 8.40  \\ 
    
    Normal & 1.54  & 1.0    & 0.59  & 8.40 &    -   \\ 
    
    \end{tabularx}
    \caption{\textit{Gradient similarity (GS).} Cosine similarity between task-specific gradient updates on the MTL backbone. Averaged across all training epochs. Multiplied by $100$.}
    \label{appendix:aff_metric_gs}
\end{table}

\begin{table}[h!]    
\small
\centering
\setlength{\tabcolsep}{3pt}
\newcolumntype{Y}{>{\centering\arraybackslash}X}
    \begin{tabularx}{\linewidth}{
    Y|
    Y
    Y
    Y
    Y
    Y
    }
    
    \multicolumn{1}{c|}{} &
    \multicolumn{5}{c}{\textbf{Affinities estimations}} \\
    
    \multicolumn{1}{c|}{with} 
    & \multicolumn{1}{c}{\textbf{SemSeg}} 
    & \multicolumn{1}{c}{\textbf{Keypts}} 
    & \multicolumn{1}{c}{\textbf{Edges}} 
    & \multicolumn{1}{c}{\textbf{Depth}} 
    & \multicolumn{1}{c}{\textbf{Normal}}  \\ 

    \hline
    
    SemSeg &   -   & +0.02    & +0.25  & +1.69 & +0.74 \\ 
     
    Keypts & -0.03  &   -   & +0.38  & -0.01 & +0.01  \\ 
    
    Edges  & -0.20  & +0.71   &   -   & +0.19 & +0.27 \\ 
    
    Depth  & +0.47  & +0.01   & +0.15 &   -   & +0.90  \\ 
    
    Normal & +0.27  & +0.03   & +0.16 & +1.26 &   -   \\ 
    
    \end{tabularx}
    \caption{\textit{Gradient transference (GT).} Look-ahead ratio simulating the effect of applying task-specific updates to the MTL backbone for the other task. Averaged across all training epochs.}
    \label{appendix:aff_metric_gt}
\end{table}

~\\
~\\
~\\
~\\
~\\
~\\
~\\
~\\
~\\
~\\
~\\

\clearpage

\section{Taskonomy buildings used} \label{appendix:splits}
We split our subset of the Taskonomy dataset into train, validation and test sets, on a per-building basis.

~\\
\textbf{Train set} These buildings amount to 603,437 input images.\\

\begin{itemize}
    \item adairsville
    \item airport
    \item albertville
    \item anaheim
    \item ancor
    \item andover
    \item annona
    \item arkansaw
    \item athens
    \item bautista
    \item bohemia
    \item bonesteel
    \item bonnie
    \item broseley
    \item browntown
    \item byers
    \item scioto
    \item nuevo
    \item goodfield
    \item donaldson
    \item hanson
    \item merom
    \item klickitat
    \item onaga
    \item leonardo
    \item marstons
    \item newfields
    \item pinesdale
    \item lakeville
    \item cosmos
    \item benevolence
    \item pomaria
    \item tolstoy
    \item shelbyville
    \item allensville
    \item wainscott
    \item beechwood
    \item coffeen
    \item stockman
    \item hiteman
    \item woodbine
    \item lindenwood
    \item forkland
    \item mifflinburg
    \item ranchester
    \item springerville
    \item swisshome
    \item westfield
    \item willow
    \item winooski
    \item hainesburg
    \item irvine
    \item pearce
    \item thrall
    \item tilghmanton
    \item uvalda
    \item sugarville
    \item silas
\end{itemize}

~\\
\textbf{Validation set} These buildings amount to 82,345 input images.\\
\begin{itemize}
    \item corozal
\item collierville
\item markleeville
\item darden
\item chilhowie
\item churchton
\item cauthron
\item cousins
\item timberon
\item wando
\end{itemize}

~\\
\textbf{Test set} These buildings amount to 40,367 input images.\\
\begin{itemize}
    \item ihlen
\item muleshoe
\item noxapater
\item mcdade
\end{itemize}

\section{Affinity scores computation}\label{appendix:aff_metrics}

\newcolumntype{Y}{>{\centering\arraybackslash}X}

\begin{table*}[h!]
    \small

    \begin{tabularx}{\textwidth}{
        >{\Centering \hsize=.12\hsize}Y| 
        >{\Centering \hsize=.06\hsize}Y| 
        >{\Centering \hsize=.55\hsize}X| 
        >{\Centering \hsize=.20\hsize}X| 
        >{\Centering \hsize=.1\hsize}X 
    }
    
    \textbf{Affinity scoring} & 
    \textbf{Type} & 
    \textbf{Computation} & 
    \textbf{Comment} & 
    \textbf{Range} \\
    
    \hline
    \hline
    
    Taxonomical distance (TD) & 
    Model-agnostic & 
    Distance between tasks in a taxonomy tree.
    &
    Symmetric. Taxonomy borrowed from \cite{zamir2018taskonomy}. Multiplied by $-1$ for consistency \ie{} higher is better. &
    $]-\infty, 0]$
    \\

     \hline
     \hline
     
    Input attribution similarity (IAS) & 
    STL-based & 
    \begin{equation}
        \frac{1}{|\mathcal{X}|}\sum_{x \in \mathcal{X}}^{} S_{cos}(Attr(STL_a, x), Attr(STL_b, x)),
    \end{equation}
    where $S_{cos}$ is cosine similarity, $\mathcal{X}$ denotes a batch of examples and $Attr$ the attribution method used.
    &
    Symmetric. Revisited from \cite{input_attr_aff_metric_ref}. Computed on a subset of the test set (2,048 images) using $InputXGradient$ attribution \cite{inputXgradient}. &
    $[-1, +1]$
    \\
     
    \hline
    Representation similarity analysis (RSA) & 
    STL-based & 
    \begin{equation}
        RSA(\theta_{Ba}, \theta_{Bb}, \mathcal{X}),
    \end{equation}
    where $RSA$ denotes the Representation Similiarity Analysis, $\mathcal{X}$ a batch of examples, $\theta_{Ba}$ and $\theta_{Bb}$ the backbone weights of the STL models for tasks $a$ and $b$ resp.
    &
    Symmetric. Revisited from \cite{TL_RSA}. Computed on a subset of the test set (2,048 images). &
    $[-1, +1]$
    \\
    
    \hline
    Label injection (LI) & 
    STL-based & 
    \begin{equation}
        \frac{\mathcal{L}_{STL_a} - \mathcal{L}_{STL_{a\leftarrow b}}}{\mathcal{L}_{STL_{a\leftarrow b}}},
    \end{equation}
    where $STL_{a\leftarrow b}$ represents the STL model for task $a$, modified to ingest the corresponding label from task $b$ in addition to the input.
    &
    Asymmetric. Novel proposal. Computed using test losses. &
    $]-\infty, +\infty[$
    
    \\
     
     \hline
     \hline
     
    Gradient similarity (GS) & 
    MTL-based & 
    \begin{equation}
        \frac{1}{N}
        \sum_{i=1}^{N}
        S_{cos}(\frac{\partial{\mathcal{L}_a(\mathcal{X},\theta_B^{i}, \theta_{Ha}^{i})}}{\partial{\theta_{B}^{i}}}, \frac{\partial{\mathcal{L}_b(\mathcal{X},\theta_B^{i}, \theta_{Hb}^{i})}}{\partial{\theta_{B}^{i}}}),
    \end{equation}
    where $N$ denotes the number of training epochs, $S_{cos}$ the cosine similarity, $\mathcal{X}$ a batch of examples, $\theta_{B}^i$ the weights of the common MTL backbone at the $i^{th}$ epoch, $\theta_{Ha}^i$ and $\theta_{Hb}^i$ the weights of the heads for $a$ and $b$ at the $i^{th}$ epoch.
    &
    Symmetric. Borrowed from \cite{fifty2021efficiently, grads_clashes}.
    &
    $[-1, +1]$\\ 
    
    \hline
     
    Gradient transference (GT) & 
    MTL-based & 
    \begin{equation}
        \frac{1}{N}\sum_{i=1}^{N}1 -
        \frac{\mathcal{L}_a(\mathcal{X},\theta_{B|b}^{i+1}, \theta_{Ha}^{i})}{\mathcal{L}_a(\mathcal{X},\theta_{B}^{i}, \theta_{Ha}^{i})},
    \end{equation}
    where $N$ denotes the number of training epochs, $\mathcal{X}$ a batch of examples, $\theta_{B|b}^{i+1}$ the weights of the common MTL backbone updated using the loss of task $b$ at the epoch $i+1$, $\theta_{Ha}^i$ and $\theta_{Hb}^i$ the weights of the heads for $a$ and $b$ at the $i^{th}$ epoch.
    &
    Asymmetric. Borrowed from \cite{fifty2021efficiently}.
    &
    $]-\infty, +\infty[$
     
    \end{tabularx}
    
    \caption{Tasks affinity scores description and computation considering two tasks $t_1 = a$ and $t_2 = b$.}
    
    \label{affinity_metrics}
  
\end{table*}

In Table \ref{affinity_metrics}, we detail the computation of each selected affinity score. Considering two tasks $t_1=a$ and $t_2=b$ and a batch of examples $\mathcal{X}$, we denote: 

\begin{itemize}

    \item their resp. losses functions $\mathcal{L}_a$ and $\mathcal{L}_b$
    \item their resp. STL models $STL_a$ and $STL_b$ with losses 
    \begin{itemize}
     \item $\mathcal{L}_{STL_a}=\mathcal{L}_a(\mathcal{X}, STL_a)$
     \item$\mathcal{L}_{STL_b}=\mathcal{L}_b(\mathcal{X}, STL_b)$
    \end{itemize}
    \item their joint MTL model $MTL_{(a,b)}$ with loss \\ $\mathcal{L}_{MTL_{(a,b)}}=\mathcal{L}_a(\mathcal{X}, MTL_{(a,b)}) + \mathcal{L}_b(\mathcal{X}, MTL_{(a,b)})$
    
\end{itemize}

Note that if the score is symmetric, it assesses how much the two tasks help each other regardless of direction. If it is asymmetric, it considers how much the target task $a$ benefits from being learned with the partner task $b$. While all scores could not be constrained to lie in the same range, higher always means more affinity.

\end{document}